\documentclass{article}

\usepackage{arxiv}

\usepackage[utf8]{inputenc} 
\usepackage[T1]{fontenc}    
\usepackage{hyperref}       
\usepackage{url}            
\usepackage{booktabs}       
\usepackage{amsfonts}       
\usepackage{nicefrac}       
\usepackage{microtype}      
\usepackage{lipsum}		
\usepackage{graphicx}
\usepackage{natbib}
\usepackage{doi}

\usepackage{rotating}
\usepackage{cleveref}
\usepackage{wrapfig}
\usepackage{graphicx}
\usepackage{bm}
\usepackage{subcaption}

\title{Distance Metric Learning Loss Functions in Few-Shot Scenarios of Supervised Language Models Fine-Tuning}

\author{ \href{https://orcid.org/0000-0002-2241-9588}{\includegraphics[scale=0.06]{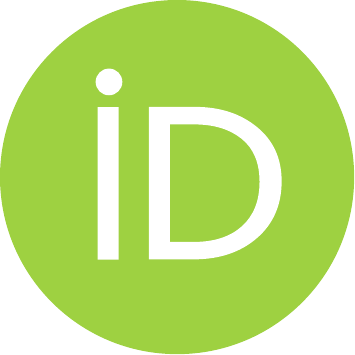}\hspace{1mm}Witold Sosnowski} \\
	Faculty of Mathematics and Information Science, \\
	Warsaw University of Technology, Poland \\
	\And
	\href{https://orcid.org/0000-0003-0617-7301}{\includegraphics[scale=0.06]{orcid.pdf}\hspace{1mm}Karolina Seweryn} \\
	Faculty of Mathematics and Information Science, \\
	Warsaw University of Technology, Poland \\
	NASK - National Research Institute, Poland \\
	\texttt{karolina.seweryn@pw.edu.pl} \\
		\And
	\href{https://orcid.org/0000-0002-3407-7570}{\includegraphics[scale=0.06]{orcid.pdf}\hspace{1mm}Anna Wróblewska} \\
	Faculty of Mathematics and Information Science, \\
	Warsaw University of Technology, Poland \\
	\texttt{anna.wroblewska1@pw.edu.pl} \\
		\And
	\href{https://orcid.org/0000-0002-9647-6761}{\includegraphics[scale=0.06]{orcid.pdf}\hspace{1mm}Piotr Gawrysiak} \\
	Faculty of Electronics and Information Technology \\
	Warsaw University of Technology, Poland \\
}

\renewcommand{\shorttitle}{Distance Metric Learning Loss Functions in Few-Shot Scenarios}

\hypersetup{
pdftitle={Distance Metric Learning Loss Functions in Few-Shot Scenarios of Supervised Language Models Fine-Tuning},
pdfauthor={Witold Sosnowski, Karolina Seweryn, Anna Wróblewska, Piotr Gawrysiak},
pdfkeywords={natural language processing, deep learning, loss functions, text classification, explainability, few-shot learning},
}

\begin{document}
\maketitle

\begin{abstract}
	This paper presents an analysis regarding an influence of the Distance Metric Learning (DML) loss functions on the supervised fine-tuning of the language models for classification tasks. We experimented with known datasets from SentEval Transfer Tasks.
  
    Our experiments show that applying the DML loss function can increase performance on downstream classification tasks of RoBERTa-large models in few-shot scenarios. Models fine-tuned with the use of \textit{SoftTriple} loss can achieve better results than models with a standard \textit{categorical cross-entropy} loss function by about 2.89 percentage points from 0.04 to 13.48 percentage points depending on the training dataset. Additionally, we accomplished a comprehensive analysis with explainability techniques to assess the models' reliability and explain their results.
\end{abstract}

\keywords{natural language processing \and deep learning \and loss functions \and text classification \and explainability \and few-shot learning}

\section{Introduction}

The development of new techniques in the Natural Language Processing (NLP) field has been studied over the last few years. It resulted in a few breakthroughs, which constantly stimulated waves of interest in the text processing area. The most recent 
discoveries are based on the 
Transformer architecture that enabled capturing the semantic meaning of the sentence~\cite{vaswani2017attention}. The Transformer architecture facilitated the discovery of the BERT encoder~\cite{devlin2018bert} that nowadays is massively used to solve most NLP tasks by adapting it in the fine-tuning process.

Unfortunately, fine-tuning of pre-trained models has a number of flaws. First of all, it is not designed to perform well when the number of observations is limited, so large training sets are consistently required for models to perform well~\cite{bansal2019learning}. Secondly, the fine-tuning process happens to be very unstable across different runs with different seeds, even though just a few minor components of the learning process are dependent on random seeds~\cite{zhang2020revisiting}. The lack of stability is even more exacerbated in the case of few-shot learning scenarios.

On the other hand, pre-trained models are fine-tuned for a specific task using a \textit{cross-entropy} objective function that focuses on learning class-specific features rather than their class representations. In other words, it only encourages inter-class distances and is not taking care of minimizing intra-class distances that would result in learning discriminative features~\cite{wen2016discriminative}. This leads to a poor generalization of the model and thus causes problems with noisy or outlier data~\cite{cao2019learning}. 

However, a few research studies proposed other loss functions, i.e. Distance Metric Learning (DML) function family, which addresses the problems of the \textit{cross-entropy} loss. The DML losses are meant to push representations of observations together if they belong to the same class and separate those from different classes. We believe this will be particularly helpful for few-shot learning, where the number of observations is insufficient for a suitable ordering of the embedding space. Therefore, we decided to investigate the effect of using loss functions from the DML family on fine-tuning the BERT-based encoder for downstream tasks in few-shot learning scenarios.

Our main contributions are the following:
\begin{enumerate}
    \item We apply \textit{SoftTriple} loss in the few-shot scenarios for the supervised fine-tuning language model in NLP domain.
    \item We examine the effect of using loss functions from the DML family on the supervised fine-tuning of the RoBERTa-large language model.
    \item We establish that \textit{SoftTriple} loss is more efficient than Supervised Contrastive loss for the supervised fine-tuning RoBERTa-large language model.
    \item After thorough experiments, our finding is that applying the DML loss to the RoBERTa-large language model is more viable the smaller the training set.
    \item Additionally, to deeper understand the models and check their reliability we propose a comprehensive analysis and explainability techniques.
\end{enumerate}

The following section is dedicated to a brief overview of the DML methods. Our new method is outlined in Section~\ref{sec:method}. The next section provides a performance analysis of the models and investigates their behaviour. Finally, a summary of the experiments is described in the last section.

\section{Methodological Background}

The DML loss can be applied whenever embedded representations of input observations are learned~\cite{qian2019softtriple}. It is most often utilized in image processing, where the high-quality image embeddings are formed into the downstream tasks, such as k-nearest neighbours classification~\cite{weinberger2009distance}, clustering or image retrieval~\cite{xing2002distance}. The main goal of DML is to minimize distances between observations from the same class while maximizing inter-class distances. Standard methods for DML is Contrastive Loss and Triplet Loss~\cite{hadsell2006dimensionality,schroff2015facenet}. In our experiments, we utilized most modern approaches such as Supervised Contrastive Learning Loss~\cite{pereyra2017regularizing} and SoftTriple Loss~\cite{qian2019softtriple}.






%


\subsection{Supervised Contrastive Learning Loss}
Supervised Contrastive Learning (\textit{SupCon}) is one of the modern DML approaches which outperforms traditional methods such as Triplet Loss or Contrastive Loss. The \textit{SupCon} introduces temperature regularization~\cite{pereyra2017regularizing} and batch processing which means that the loss is not calculated just for a single triplet but is the average of all possible triplets from a given batch.
The \textit{SupCon} loss extends the self-supervised batch DML approach to the fully-supervised setting, which provides the ability to use the label information~\cite{khosla2020supervised}. Instead of contrasting one positive example for an anchor with all other observations from the batch, SupCon contrasts all examples from the same class (as positives) with all other observations from the batch as negatives. The most critical issue with this approach is that as the number of observations in the batch grows, the number of triplets grows cubically.

The SupCon loss has been previously applied in the supervised fine-tuning of the RoBERTa-large language model~\cite{gunel2020supervised} and its formula is as follows:
\begin{equation} \label{eq:supcon-loss}
\resizebox{0.8\textwidth}{!}{
$\ell_{SupCon}=\sum_{i \in I} \frac{-1}{|P(i)|} \sum_{p \in P(i)} \log \frac{\exp \left(\boldsymbol{z}_{i} \cdot \boldsymbol{z}_{p} / \tau\right)}{\sum_{a \in A(i)} \exp \left(\boldsymbol{z}_{i} \cdot \boldsymbol{z}_{n} / \tau\right)},$
}
\end{equation}

\noindent where $i \in I \equiv\{1, \ldots, N\}$ denotes the index of an anchor observation $x_i$, $P(i)$ is the set of observations from the same class that $x_i$ belongs to, $A(i) \equiv I \backslash\{i\}$, $z_{i}$ denotes the representation of the $x_{i}$ observation that is outputted by the encoder. 
$x_{p}$ denotes observation from the same class as $x_i$ (positive class), while $x_{n}$ denotes the observations from different class as $i$. $\tau \in \mathcal{R}^{+}$ is a scalar temperature. 

\subsection{SoftTriple Loss}

\textit{SoftTriple} loss is another modern DML approach that extends the batch supervised DML approach. It employs proxies -- artificial embeddings that are learned during the training procedure  -- to represent classes from the dataset. In other words, each class has its own additional set of points that best approximates the position of all observations from the class. The number of proxies is given as a hyperparameter and should reflect the global distribution of features from the class, which may have more than one focal point.   
The loss is calculated based on the possible triplets in the batch. It is the same as the \textit{SupCon} loss. The difference is in calculating the single triplet. It is composed of the anchor (observation), positive proxy (a proxy from the anchor class) and negative proxy (a proxy from a class other than the anchor).
The number of all possible triples is linear with respect to the number of original examples, while it grows cubically (as stated earlier) for the \textit{SupCon} or other non-proxy losses. Therefore, this approach consumes much fewer resources and has higher performance because the proxy provides better resistance to outliers. 

The following formulas define the \textit{SoftTriple} loss:
\begin{equation} \label{eq:softtriple-loss}
\resizebox{0.8\textwidth}{!}{
$\ell_{SoftTriple}
=-\frac{1}{N} \sum_{i \in I}{\log \frac{\exp \left(\lambda\left(\mathcal{S}_{i, y_{i}}^{\prime}-\delta\right)\right)}{\exp \left(\lambda\left(\mathcal{S}_{i, y_{i}}^{\prime}-\delta\right)\right)+\sum_{j \neq y_{i}} \exp \left(\lambda \mathcal{S}_{i, j}^{\prime}\right)}}$
}
\end{equation}

\begin{equation} \label{eq:softtriple-loss2}
\mathcal{S}_{i, c}^{\prime}=\sum_{k \in K} \frac{\exp \left(\frac{1}{\gamma} {E({x}_{i})}^{\top} {w}_{c}^{k}\right)}{\sum_{k \in K} \exp \left(\frac{1}{\gamma} {E({x}_{i})}^{\top} {w}_{c}^{k}\right)} {E({x}_{i})}^{\top} {w}_{c}^{k},
\end{equation}

\noindent where $i \in I \equiv\{1 \ldots N\}$ denotes the index of an anchor observation $x_i$, $y_i$ its corresponding label, $c \in C$ denotes the class index, $C$ denotes the class number, $k \in K$ denotes the proxy index of class $c$, $K$ denotes the number of proxies per class, $\delta$ denotes a minimum interclass margins, $\lambda$ scaling factor that reduces the outliers' impact, $\gamma$ scaling factor for the entropy regularizer,  $x_{i}$ denotes a $i$th observation' embedding, $E(\cdot) \in \mathbf{R}^{d}$ denotes an encoder and ${w}_{c}^{k}$ are weights that represent embeddings of the class $c$.

\section{Our Approach}
\label{sec:method}
We analyze the impact of the DML loss function on the supervised training of the pre-trained language model by extending the loss function with \textit{SupCon} loss or \textit{SoftTriple} loss. In general, the new loss function contains two elements; the first one is \textit{categorical cross-entropy} loss applied to the output of the classification layer, and the DML loss is applied directly to the output of the encoder. The second part of our loss function maximizes distances between observations from different classes and minimizes distances between observations from the same class.



The goal function is given by the following formula:
\begin{equation}
\mathcal{L}=(\beta) \mathcal{\ell}_{CCE}+(1 - \beta) \mathcal{\ell}_{DML},
\mathcal\ell_{CCE}=-\frac{1}{N} \sum_{i=1}^{N} \sum_{c=1}^{C} y_{i, c} \cdot \log \hat{y}_{i, c}
\label{eq:novel-loss}
\end{equation}
where
$N$ is the number of the observations,
$C$ denotes the class number,
$y_{i c}$ denotes the label of the class $c$ for the $i$th observation,
$p_{i c}$ denotes the models's predicted probability of the $i$th observation for the $c$th class
$\beta$ denotes the scaling factor that tunes influence of both parts of the loss,
$\ell_{{DML }}$ denotes either $\ell_{SupCon}$ or $\ell_{SofTriple}$.
$\ell_{{CCE }}$ is the \textit{categorical cross-entropy}.

\subsection{Model Architecture}

We compare the loss functions for supervised fine-tuning of the RoBERTa-large~\cite{liu2019roberta} encoder provided by the \textit{huggingface} library as the pre-trained model \textit{roberta-large}. The standard supervised fine-tuning procedure starts by input text tokenization with the use of WordPiece tokenizer~\cite{wu2016google}. The tokenized text is represented as an array of tokens with the \textit{[CLS]} token at the beginning, the \textit{[EOS]} at the end and the \textit{[SEP]} separating sentences. The array of tokens is then passed to the RoBERTa model $E(\cdot)$. The encoder output is an array of embeddings corresponding to the array of input tokens. In the standard setup, the embedding array is passed to a fully connected layer from which as many neurons as we have classes are output. The output of this layer is then passed to the softmax function and then to the \textit{categorical cross-entropy} function, which, while fed with labels, calculates the loss.

Our experiments involve modifying the loss function by adding the part calculated from the DML loss. In particular, we analyze the effect of introducing an additional \textit{SupCon} loss or a \textit{SoftTriple} loss.

\subsubsection{CCE + SupCon.}
The first experimental setting includes fine-tuning a model operating on the loss function consisting of the \textit{categorical cross-entropy} loss and \textit{SupCon} loss according to Formula~\ref{eq:novel-loss}. The \textit{SupCon} objective calculates the loss on the $x_i$ representation outputted by the encoder $E(\cdot)$. In our case, the encoder is RoBERTa-large, and a single observation is represented as the vector associated with the \textit{[CLS]} token. A single loss is calculated based on the $x_i$ representations from the batch. Such architecture was previously proposed by~\cite{gunel2020supervised} and outperformed a baseline in a few-shot learning setting.

\subsubsection{CCE + SoftTriple.}
The second experimental setting consists of the \textit{SoftTriple} loss function built according to Formula~\ref{eq:novel-loss}, where the $\ell_{{DML }}$ part is represented by the $\ell_{SofTriple}$. Similarly, as in the case of \textit{SupCon} loss, the \textit{SoftTriple} objective calculates the loss based on the batch of $x_i$ representations outputted by the encoder $E(\cdot)$. Here we also use RoBERTa-large as the encoder, but instead of treating a single \textit{[CLS]} vector as the observation representation, we use the entire encoder output. Such implementation can be considered as the development of the idea where the \textit{SupCon} loss was applied only to the representation of the \textit{[CLS]} token~\cite{gunel2020supervised}. Feeding the DML loss to the entire model output may provide better embedding generalization but is less resource-efficient. The \textit{SoftTriple} loss uses a proxy to compute the loss, which reduces and balances the resource requirements of the entire procedure. This is the first time, this architecture is studied in the few-shot learning settings.

\subsection{Training Procedure}

We conducted experiments operating on 40-fold cross-validation; each fold was run 40 times. Thus, each result is an average F1-score of 40 runs from 40-fold cross-validation. The training was done in few-shot learning scenarios, so the required training set sizes were limited to 20 and 100 observations. We also conducted experiments on datasets limited to 1,000 observations, for comparison purposes. The 40-fold cross-validation was introduced to tackle the problem of high variance, which is natural in the few-shot learning settings~\cite{dvornik2019diversity}. 

For each task, the baseline result was obtained separately based on the hyperparameter sweep with the same batch size $=64$, learning rates $\in$\{$1e-5$, $2e-5$, $3e-5$\}, epochs number $\in\{8,16,64,128\}$, linear warmup for the first 6\% of steps and weight decay coefficient $=0.01$.
The best hyperparameter set for each task includes a learning rate of $1e-5$ and $8$ epochs for 1,000 elements datasets, $64$ for 100-element datasets and $128$ for 20-element datasets.

\begin{table}[ht]
    \parbox{.45\linewidth}{
        \centering
            \caption{\textit{SoftTriple} hyperparameters.}
            \label{tab:hyperparameters-softtriple}
            \begin{tabular}{|l|l|l|l|l|l|l|}
            \hline
            \textbf{Loss} & \textbf{\# train} & \textbf{k} & \boldsymbol{$\gamma$} & \boldsymbol{$\lambda$} & \boldsymbol{$\delta$} & \boldsymbol{$\beta$} \\
            \hline
            SoftTriple & 20 & $25$ & $0.1$ & $9$ & $0.7$ & $0.4$ \\
            SoftTriple & 100 & $2000$ & $0.1$ & $4$ & $0.7$ & $0.8$ \\
            SoftTriple & 1,000 & $2000$ & $0.1$ & $7$ & $0.9$ & $0.9$ \\
            \hline 
            \end{tabular}
    }
    \hfill
    \parbox{.45\linewidth}{
        \centering
                \caption{\textit{SupCon} hyperparameters. \label{tab:hyperparameters-supcon}}
            \begin{tabular}{|l|l|l|l|}
            \hline
            \textbf{Loss} & \textbf{\# train} & \boldsymbol{$\beta$} & \boldsymbol{$\tau$} \\
            \hline
            SupCon & 20 & $0.9$ & $0.6$ \\
            SupCon & 100 & $0.9$ & $0.7$ \\
            SupCon & 1,000 & $0.9$ & $0.7$ \\
            \hline 
            \end{tabular}
}
\end{table}

The hyperparameter search for models with the DML loss function includes additional elements. For the \textit{SupCon} loss, the grid search was performed on the following parameters: $\tau \in\{0.1, 0.3, 0.5, 0.7, 0.9\}$ and $\beta \in\{0.1, 0.3, 0.5, 0.7, 0.9\}$. In the case of \textit{SoftTriple} loss, we searched the following parameters: $k \in\{5, 25, 1000, 2000\}$, $\gamma \in\{0.01, 0.03, 0.05, 0.07, 0.1\}$, $\lambda \in\{1,3,3.3,4,6,8,10\}$, $\delta \in\{0.1, 0.3, 0.5, 0.7, 0.9, 1\}$ and $\beta \in\{0.1, 0.3, 0.5, 0.7, 0.9\}$. 
In most cases, the same hyperperameter set was present across datasets with the same size and for the \textit{SupCon} it is shown in the Table~\ref{tab:hyperparameters-supcon} while for the \textit{SofTriple} it is refered in the Table~\ref{tab:hyperparameters-softtriple}.


\subsection{Datasets}

\begin{sidewaystable}
\begin{center}
\begin{minipage}{\textheight}    

\caption{F1 score of RoBERTa-large (RL) vs RoBERTa-large with Supervised Contrastive Learning (\textit{SupCon}) loss vs RoBERTa-large with \textit{SoftTriple} loss in the few-shot learning scenarios. N means number of observations.}

\vspace*{1 cm}

\subcaption{RL vs RL-SupCon vs RL-SoftTriple trained on the 20 element datasets. \label{tab:supcon-softriple-20-examples-dataset}}

\begin{tabular}{|l|l|c|l|l|l|l|l|l|l|l|}
\hline
\textbf{Model} & \textbf{Loss} & \textbf{N} &      \textbf{SST2}                  & \textbf{MR}                  & \textbf{MPQA}         & \textbf{SUBJ}         & \textbf{TREC}         & \textbf{CR}           & \textbf{MRPC} & \textbf{Avg}\\
\hline
RoBERTa-large & CCE &   20 & 53.71$\pm$8.74          & 56.55$\pm$8.67 & 65.66$\pm$5.01 & 85.54$\pm$6.38 & 41.48$\pm$9.46 & 65.03$\pm$8.26 & 63.30$\pm$7.71 & 61.61\\
RoBERTa-large & CCE + SupCon  & 20 & 60.04$\pm$8.98          & 65.74$\pm$9.69 & 64.92$\pm$4.91 & 87.66$\pm$4.76 &    42.81$\pm$10.55        & 66.59$\pm$6.64 & 68.85$\pm$5.42 & 	65.23\\
\multicolumn{3}{|c|}{p-value} &0.002 & $<$0.001 & 0.5&0.096 & 0.555&0.355 &$<$0.001 &\\
\hline
RoBERTa-large & CCE + SoftTriple  & 20 &\textbf{62.35}$\pm$7.44 & \textbf{70.03}$\pm$8.16 & \textbf{67.96}$\pm$4.72 & \textbf{89.48}$\pm$4.62 & \textbf{47.21}$\pm$9.44 & \textbf{68.55}$\pm$6.91 & \textbf{69.32}$\pm$4.71&\textbf{67.84} \\
\multicolumn{3}{|c|}{p-value} &$<$0.001 & $<$0.001&0.038 & 0.002&0.008 & 0,042& $<$0.001&\\
\hline
\end{tabular}

\vspace*{1 cm}

\subcaption{RL vs RL-SupCon vs RL-SoftTriple trained on the 100 element datasets. \label{tab:supcon-softriple-100-examples-dataset}}

\begin{tabular}{|l|l|c|l|l|l|l|l|l|l|l|}
\hline
\textbf{Model} & \textbf{Loss} & \textbf{N} &      \textbf{SST2}                  & \textbf{MR}                  & \textbf{MPQA}         & \textbf{SUBJ}         & \textbf{TREC}         & \textbf{CR}           & \textbf{MRPC} & \textbf{Avg}\\
\hline
RoBERTa-large & CCE               & 100 &85.87$\pm$5.50          & 82.57$\pm$5.94       & 82.50$\pm$5.65  & 93.91$\pm$1.47 & 82.72$\pm$4.73 & 89.43$\pm$3.65  & 72.13$\pm$4.24   & 84.16\\
RoBERTa-large & CCE + SupCon      & 100 &\textbf{88.47}$\pm$1.38  &85.24$\pm$3.30      & 81.18$\pm$6.16   & 94.39$\pm$1.43  & 81.82$\pm$4.51 & 90.73$\pm$3.02   &72.85$\pm$4.41   &  84.95\\
\multicolumn{3}{|c|}{p-value} &0.006 & 0.016 & 0.321&0.143 & 0.386&0.087 &0.459 &\\
\hline
RoBERTa-large & CCE + SoftTriple  & 100 &88.12$\pm$1.83       & \textbf{85.49}$\pm$3.14 & \textbf{85.26}$\pm$4.12 & \textbf{94.57}$\pm$1.39 & \textbf{83.51}$\pm$4.37 & \textbf{91.16}$\pm$3.41 & \textbf{73.95}$\pm$4.34&\textbf{86.01} \\
\multicolumn{3}{|c|}{p-value} &0.018 & 0.008&0.015 & 0.042&0.44 & 0.031& 0.062&\\
\hline
\end{tabular}

\vspace*{1 cm}

\subcaption{RL vs RL-SupCon vs RL-SoftTriple trained on the 1000 element datasets}

\begin{tabular}{|l|l|c|l|l|l|l|l|l|l|l|}
\hline
\textbf{Model} & \textbf{Loss} &    \textbf{N} &                   \textbf{SST2}                  & \textbf{MR}                  & \textbf{MPQA}         & \textbf{SUBJ}         & \textbf{TREC}         & \textbf{CR}           & \textbf{MRPC} & \textbf{Avg}\\
\hline
RoBERTa-large & CCE          &1000     & 91.59$\pm$0.69          & 89.73$\pm$2.20       & 90.16$\pm$1.25  & 96.04$\pm$1.30 & 89.89$\pm$2.87 & 93.16$\pm$2.60  & 77.65$\pm$3.63   & 89.75\\
RoBERTa-large & CCE + SupCon   &1000   & \textbf{91.98}$\pm$0.70  &89.91$\pm$2.17      & 89.94$\pm$1.95     & 96.23$\pm$1.16  & \textbf{90.34}$\pm$2.66     & 93.90$\pm$2.63  &79.01$\pm$4.21   &  90.19\\
\multicolumn{3}{|c|}{p-value} &0.014 & 0.714 & 0.550&0.492 & 0.469&0.209 &0.126 &\\
\hline
RoBERTa-large & CCE + SoftTriple &1000 &\textbf{91.98}$\pm$0.81&\textbf{90.02}$\pm$2.00&\textbf{90.60}$\pm$1.74& \textbf{96.26}$\pm$1.20 & 89.93$\pm$2.45    & \textbf{94.00}$\pm$2.28 & \textbf{79.55}$\pm$4.12& \textbf{90.33} \\
\multicolumn{3}{|c|}{p-value} &0.023 & 0.539&0.198 & 0.434&0.947 & 0,129& 0.032&\\
\hline
\end{tabular}
\label{tab:supcon-softriple-20-examples-dataset}
\end{minipage}
\end{center}
\end{sidewaystable}

The datasets on which we tested the models come from the well-known SentEval Transfer Task, which includes classification and textual entailment tasks~\cite{conneau2018senteval}, giving a fair picture of overall model performance (see Table~\ref{tab:datasets_description}). 

\begin{table}[!ht]
\centering
\caption{SentEval Transfer Task datasets used for our experiments.}
\label{tab:datasets_description}
\begin{tabular}{|l|l|l|l|}
\hline
\textbf{Dataset} &\textbf{ \# Sentences} & \textbf{\# Classes} & \textbf{Task}\\
\hline
SST2 & 67k & 2 &  Sentiment (movie reviews)\cite{socher2013recursive}\\
MR & 11k & 2 &  Sentiment (movie reviews) \cite{pang2005seeing} \\
MPQA & 11k & 2 &  Opinion polarity \cite{wiebe2005annotating}\\
SUBJ & 10k & 2 &  Subjectivity status \cite{pang2004sentimental}\\
TREC & 5k & 6 &  Question-type classification \cite{pang2005seeing}\\
CR & 4k & 2 &  Sentiment (product review) \cite{hu2004mining}\\
MRPC & 4k & 2 & Paraphrase detection \cite{dolan2004unsupervised}\\
\hline 
\end{tabular}
\end{table}

\begin{table*}[h!]
\centering
\caption{Detailed metrics for few-shot learning scenario with 20 training samples from the MR dataset.}
\begin{tabular}{|l|l|l|l|l|}
\hline
\textbf{Model} & \textbf{F1 score} & \textbf{Accuracy}  & \textbf{Recall} & \textbf{Precision} \\
\hline

CCE   & 56.55 $\pm$ 8.67 & 59.98 $\pm$ 5.84 & 61.73 $\pm$ 26.19 & 63.98 $\pm$ 13.30  \\
CCE + SupCon  & 65.74 $\pm$ 9.69  & 67.22 $\pm$ 8.30  & 62.10 $\pm$ 20.48 & 71.74 $\pm$ 11.16 \\
CCE + SoftTriple & \textbf{70.03} $\pm$ 8.16 & \textbf{71.01} $\pm$ 7.08  & \textbf{71.3} $\pm$ 16.59  & \textbf{73.48} $\pm$ 10.71\\
\hline 
\end{tabular}
\label{tab:few-shot-more-metrics}
\end{table*}

\begin{figure*}[h!]
\centering
\includegraphics[width=.3\textwidth]{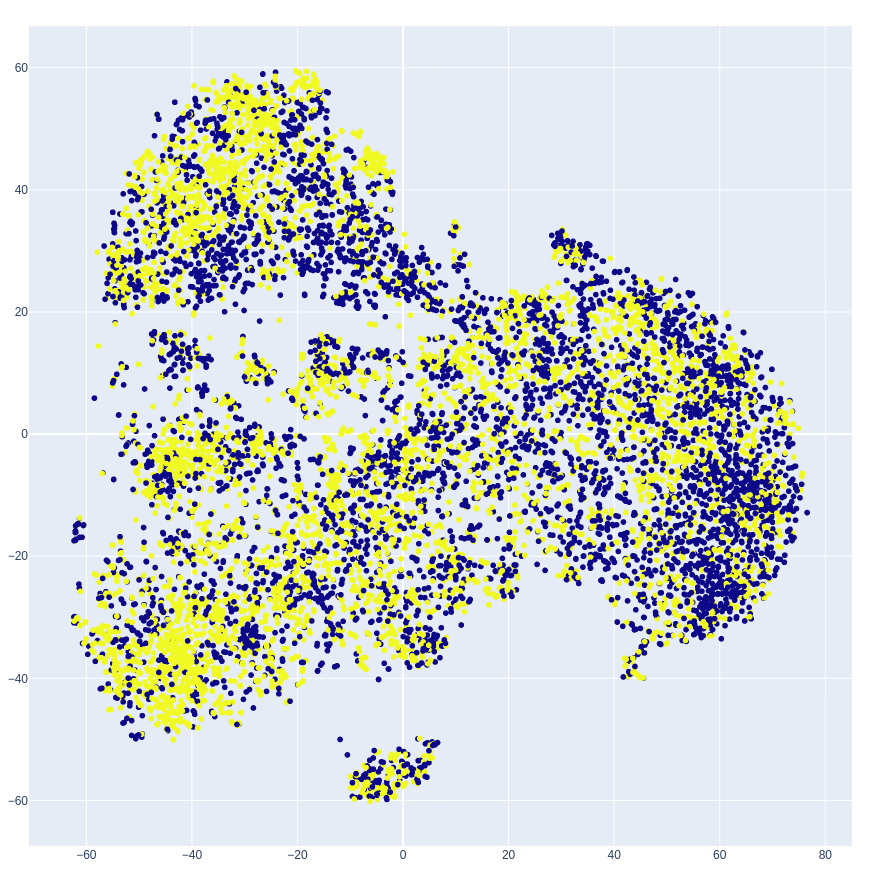}\hfill
\includegraphics[width=.3\textwidth]{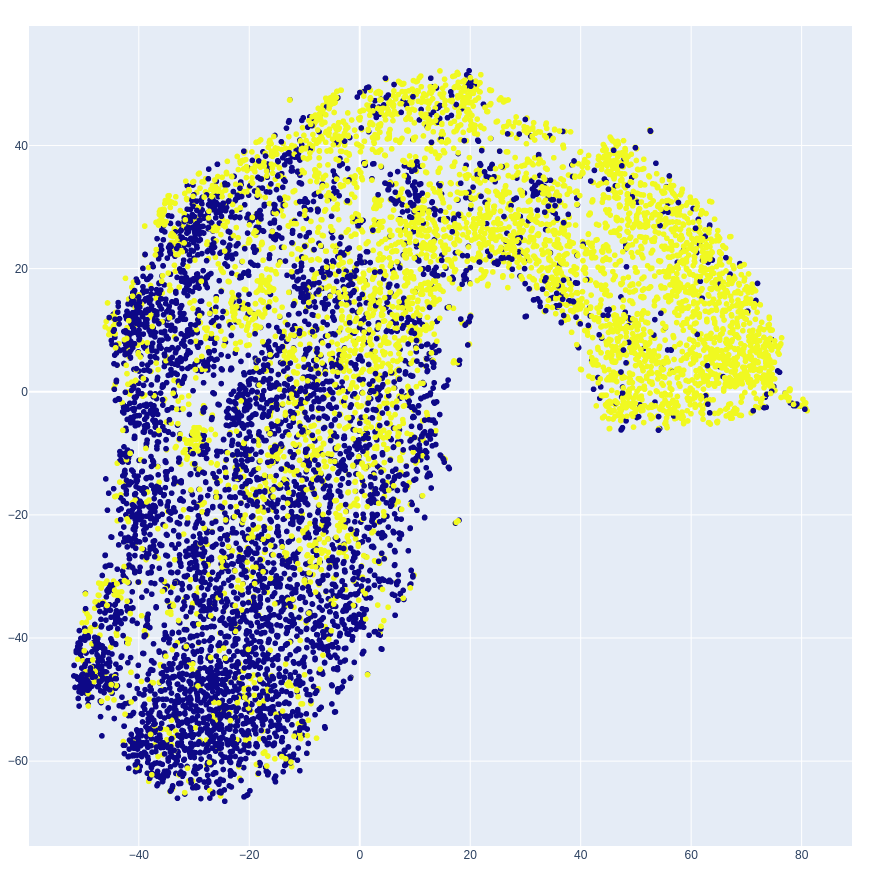}\hfill
\includegraphics[width=.3\textwidth]{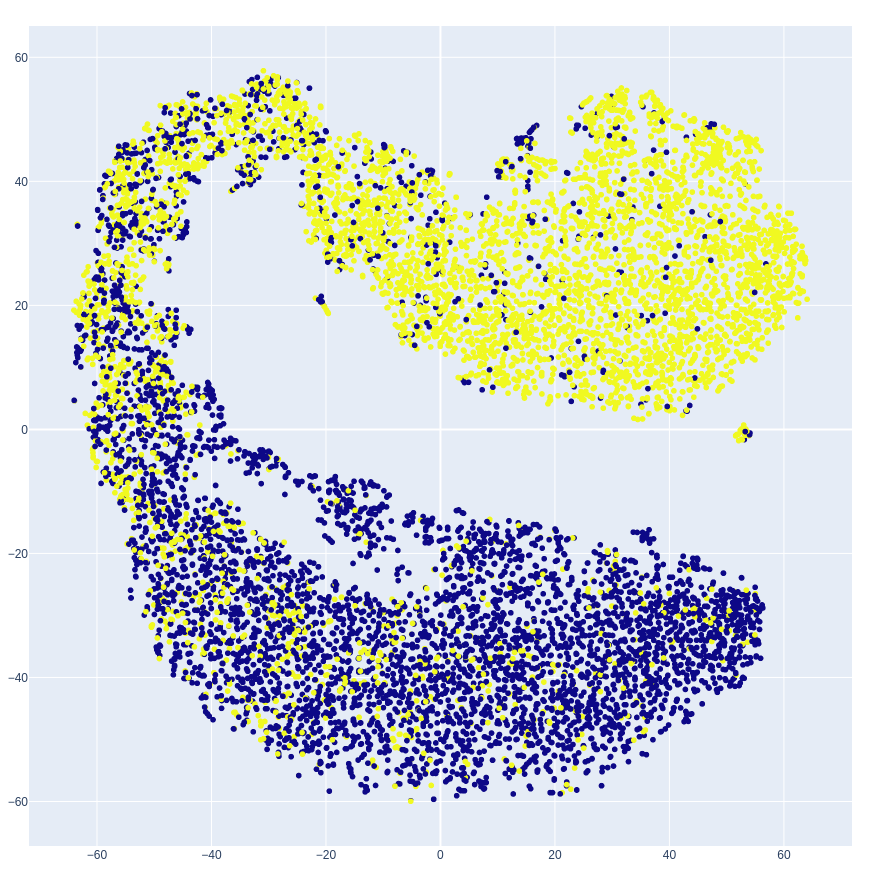}\hfill
\caption{t-SNE visualization of [CLS] embeddings for few-shot models trained on 20 examples from the MR dataset with different loss functions. Colour represents sentiment: blue - negative, and yellow - positive. Left: \textit{\textit{cross-entropy} (CCE)}, Middle: \textit{CCE + SupCon} Loss, Right: \textit{CCE + SoftTriple} Loss.}
\label{fig:tsne}
\end{figure*}

\section{Experimental Results}
\label{sec:experiments}

\subsection{Result Analysis}

\Cref{tab:supcon-softriple-20-examples-dataset} presents our results separately for models trained on 20, 100 and 1,000 observations. The results show a baseline performance where the model was trained using only \textit{CCE loss} compared to models trained on \textit{CCE + SupCon} loss and \textit{CCE + SoftTriple} loss.

\subsubsection{20-element dataset}
Models with the \textit{SoftTriple} loss outperformed baseline and models with the \textit{SupCon} loss in each case, giving an average gain of 6.23 percentage points over baseline and 2.61 percentage points over the \textit{SupCon} approach. The most significant improvement over baseline was observed for the MR dataset and the \textit{SoftTriple} loss model, almost 13.5 percentage points. On the other hand, the lowest improvement from baseline for the \textit{SoftTriple} loss models is 2.29 percentage points for the MPQA dataset, while the \textit{SupCon} showed a deterioration in performance from baseline on this dataset. 

\subsubsection{100-element dataset}

The \textit{SofTriple} loss models are better than the baseline model in every case, resulting in an average improvement of 1.85 percentage points. Moreover, the \textit{SoftTriple} loss models outperform the \textit{SupCon} loss models in every case, except for the SST2 dataset, where the \textit{SupCon} model is better than the \textit{SoftTriple} by 0.35 percentage points.  The largest improvement over baseline was observed for the MR set and the \textit{SoftTriple} loss model and is 2.91 percentage points. On the other hand, the lowest improvement from baseline for the \textit{SoftTriple} loss models is 0.67 percentage points for the SUBJ set. 

\subsubsection{1,000-element dataset}

The \textit{SofTriple} loss models are better than the baseline model in every case, resulting in an average improvement of 0.58 percentage points. Furthermore, the \textit{SoftTriple} loss models outperform the \textit{SupCon} loss models in every case, except for the TREC dataset, while in the case of SST2 the performance was the same. The largest improvement over baseline was observed for the MRPC set and the \textit{SoftTriple} loss model and is 1.9 percentage points. On the other hand, the lowest improvement from baseline for the \textit{SoftTriple} loss models is 0.67 percentage points for the SUBJ dataset. 
The average performance of the models trained using \textit{CCE} loss combined with \textit{SoftTriple} loss is higher than the baseline or models trained using \textit{CCE} and \textit{SupCon} loss.

\subsection{Comprehensive Analysis}

Among all the results, the largest percentage improvement of the F1 score compared to the  baseline (\textit{CCE} loss function) is for the MR dataset (+24\% of relative difference, +13.48 of absolute difference). This section investigates sentiment analysis models trained in the few-shot scenarios on the smallest analyzed training dataset: 20 samples from the MR dataset. 
We conducted a comprehensive investigation to compare models' results for those observations.

Table~\ref{tab:few-shot-more-metrics} summarizes various metrics proving that the proposed method (\textit{CCE + SoftTriple}) outperforms other approaches (\textit{CCE}, \textit{CCE + SupCon}) on the MR dataset. Particularly, considerable progress has been made regarding the recall. In this case, \textit{CCE + SupCon} achieves similar results to the base model (\textit{CCE}), while values for \textit{CCE + SoftTriple} differ significantly. 
Smaller values of standard deviation indicate that models with \textit{CCE + SoftTriple} are more stable.

Figure~\ref{fig:tsne} illustrates 2-dimensional representation of input text embeddings made with t-SNE algorithm~\cite{tsne}. 
The baseline model (\textit{CCE}) is not able to separate the different classes. Much better results are achieved with additional \textit{SupCon} loss. However, the observations in the middle of the plot are not decoupled. \textit{SoftTriple} loss helps to almost isolate two classes. Even 20 observations for fine-tuning the model are enough to achieve satisfactory results. 

\begin{wrapfigure}{r}{0.6\textwidth}
\centering
\vspace{-10pt}
\includegraphics[width=.6\textwidth]{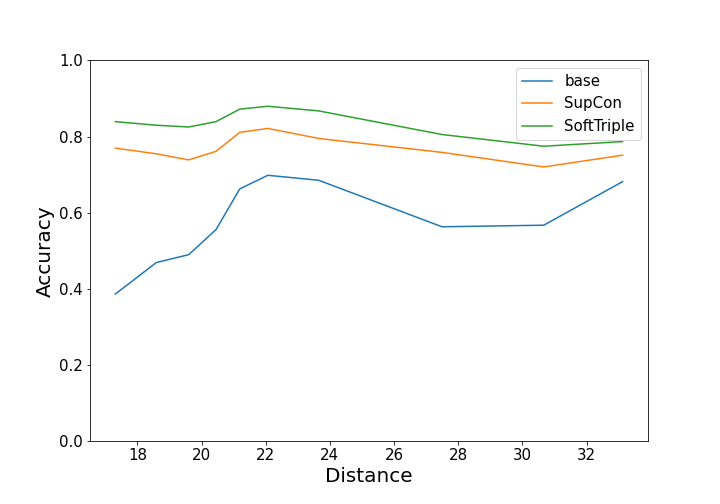}\hfill
\vspace{-20pt}
\caption{Relationship between accuracy and Euclidean distance between sentence embedding and mean embedding computed by baseline model with \textit{cross-entropy} loss.}
\label{fig:distance_accuracy}
\vspace{-10pt}
\end{wrapfigure}

Figure~\ref{fig:distance_accuracy} shows a difference in models' performance depending on the chosen loss function. in this test, we defined a mean sentence embedding based on [CLS] representation from the 
model with \textit{cross-entropy} loss. Then, we computed the Euclidean distance between the given observation and the mean embedding. The farther from the 
center, the greater the chance of 
being an outlier. Next, the observations were sorted according to the distance from the central embedding and divided into groups of equal size. The models' accuracy was calculated in the designated groups and presented on the chart in Figure~\ref{fig:distance_accuracy}. 
Both models with additional \textit{SupCon} and \textit{SoftTriple} losses achieve better results than the base model with CCE. Results of models trained using DML 
show similar trend, but the \textit{SoftTriple} is more accurate than the training with \textit{SupCon} loss, even for observations far from the central embedding.

Table~\ref{tab:examples} presents examples of explainability with Shapley values~\cite{shap}. 
In Example 1, all the models predict the correct negative sentiment for the observation. The explanations of these models indicates that the same words increase the probability of the predicted class for all analyzed models.
Intuitively, the word \textit{terrible} has a big influence on this prediction. Surprisingly, word \textit{painful} increases the probability of negative sentiment only for model with \textit{CCE + SoftTriple} loss. Other models treat this word as neutral.

In Example 2, in Table~\ref{tab:examples}, we present an observation that all analyzed models classified incorrectly as positive sentiment. Interestingly, the statement consists of word \textit{puzzling} whose sentiment can be interpreted differently depending on context. Model trained with \textit{SupCon} loss function treats this word as positive while the sentiment of this word measured with other models is slightly more negative. 

\begin{table}[ht]
\caption{A Shapley values for each token for samples from MR dataset for a model trained on 
20 samples. 
Red shades mean that a word increases the likelihood of positive sentiment, while blue shades mean 
an increase in the probability of 
a negative class.  \label{tab:examples}}
    \parbox{.6\linewidth}{
        \centering
                    \vspace{-5pt}
                    \begin{tabular}{|lr|}
                    \hline
                 \multicolumn{2}{|l|}{\textbf{Example 1}} \\
                 \hline
                    \textbf{Loss function} & \textbf{Sentiment} \\
                    \hline  
                    \textit{\textit{cross-entropy}} & positive \\
                    \multicolumn{2}{|l|}{\includegraphics[height=0.4cm]{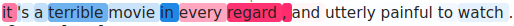}}\\
                    \hline
                    \textit{SupCon} & negative \\
                    \multicolumn{2}{|l|}{\includegraphics[height=0.4cm]{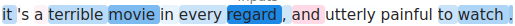}}\\
                    \hline
                    \textit{SoftTriple} & negative \\
             \multicolumn{2}{|l|}{\includegraphics[height=0.4cm]{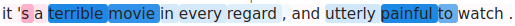}}\\
                    \hline 
                    ground-truth & negative \\
                    \hline
                    \end{tabular}
    }
    \hfill
    \parbox{.33\linewidth}{
        \centering
                \vspace{-5pt}
                \begin{tabular}{|l|r|}
                \hline
                 \multicolumn{2}{|l|}{\textbf{Example 2}} \\
                 \hline
                \textbf{Loss function~} & \textbf{Sentiment} \\
                \hline  
                \textit{\textit{cross-entropy}} & positive \\
                \multicolumn{2}{|l|}{\includegraphics[height=0.4cm]{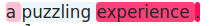}}\\
                \hline
                \textit{SupCon} & positive \\
                \multicolumn{2}{|l|}{\includegraphics[height=0.4cm]{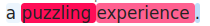}}\\
                \hline
                \textit{SoftTriple} & positive \\
                \multicolumn{2}{|l|}{\includegraphics[height=0.4cm]{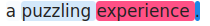}}\\
                \hline 
                ground-truth & negative \\
                \hline

                \end{tabular}
}
\end{table}

Additionally, we performed test set analysis with \textit{geval} method~\cite{gralinski-etal-2019-geval}. This methods searches features that lower the evaluation metric in which it cannot be ascribed to a chance, measured by their p-values of Mann-Whitney rank U test.
Table~\ref{tab:geval_results} shows a few groups of input observations that are particularly difficult for our models. For example, one group contains all the observations from the test set that contains the "but" word. Here, in sentiment analysis, even one word "but" can change the whole sentence sense and the model result, e.g. "Its premise is smart, but the execution is pretty weary."
Still, even for those hard groups of input examples, the \textit{SubCon} is better than the baseline (\textit{CCE}), and the \textit{SoftTriple} achieved the best performance. 

\begin{table}[!ht]
\centering
\caption{Accuracy of the three tested loss function in model training: \textit{CCE}, \textit{CCE+SubCon}, \textit{CCE+SoftTriple}, for groups of input examples with a given characteristic, e.g. all the input examples containing a word "will".}
\begin{tabular}{|c|c|c|c|c|}
\hline
\textbf{Sample group} & \textbf{\# examples} & \textbf{CCE} & \textbf{CCE+SupCon} & \textbf{CCE+SoftTriple}\\
\hline
"n't" &	884	& 52 & 71 & 78 \\
"but" &	1613	& 55 & 0.72 & 76 \\
"if" &	524	& 52 &	70 &	74 \\
"you" &	880	& 52 &	66 &	72 \\
"more" &	685 &	54 &	72 &	79 \\
"will" &	349	& 53 &	69 &	73 \\
\multicolumn{2}{|l|}{Accuracy in the whole test set} &		58 &	77 &	83 \\
\hline 
\end{tabular}

\label{tab:geval_results}
\end{table}

\section{Conclusion}

This paper investigated the impact of using DML loss functions on supervised fine tuning language models by comparing the results of RoBERTa-large trained with raw \textit{\textit{cross-entropy}} loss function and training procedure enriched with distance metrics: \textit{SupCon} and \textit{SoftTriple} loss. The performance of the models was tested based on 40-fold cross-validation over multiple datasets from the SentEval Transfer Tasks. We found that the use of these functions improves the performance of the model, and the improvement is greater the smaller the dataset is. Applying \textit{SoftTriple} loss increases the performance over baseline on average by about 2.89 percentage points, from 0.04 to 13.48 percentage points which is superior comparing to the average 1.62 percentage points from -0.22 to 9.19 percentage points for the \textit{SupCon}. We would like to acknowledge that most of the results for which \textit{p-value} is less than 0.05 belong to few-shot learning scenarios (20-element datasets, 100-element datasets), which is consistent with the results from previous work~\cite{gunel2020supervised}.

The comprehensive analysis of few-shot models' performance on sentiment analysis task from the MR dataset revealed that enriching training procedure with DML methods (both \textit{SupCon} and \textit{SoftTriple}) cause an increase in model accuracy. Model fitted with \textit{SofTriple} loss achieved higher scores for all analysed metrics (F1 score, accuracy, recall, precision, and AUC). Moreover, t-SNE visualizations point towards the idea that our method separates classes better than other models even for 20 training observations. 

\bibliographystyle{unsrtnat}
\bibliography{references}  






\end{document}